\documentclass{article}

\usepackage{arxiv}

\usepackage[utf8]{inputenc} 
\usepackage[T1]{fontenc}    
\usepackage{hyperref}       
\usepackage{url}            
\usepackage{booktabs}       
\usepackage{amsfonts}       
\usepackage{nicefrac}       
\usepackage{microtype}      
\usepackage{lipsum}		
\usepackage{graphicx}
\usepackage{amsmath}
\usepackage{doi}
\usepackage{comment}
\usepackage{caption}
\usepackage{cite}
\usepackage{subcaption}
\usepackage{authblk}
\usepackage[english]{babel}
\usepackage{csquotes}
\usepackage{pgfplots}
\pgfplotsset{compat=1.18}
\usetikzlibrary{fit}
\usepackage{float}
\usepackage{xcolor}
\definecolor{darkgreen}{rgb}{0.0, 0.5, 0.0} 
\definecolor{teal_blue}{HTML}{003B46} 
\definecolor{ash_gray}{HTML}{BDC3C7} 
\definecolor{burgundy}{HTML}{6A1B4D}
\definecolor{mustard}{HTML}{E1AD01}

\usepgfplotslibrary{colorbrewer}
\usetikzlibrary{pgfplots.statistics, pgfplots.colorbrewer} 
\usepackage{pgfplotstable}
\usepackage{filecontents}

\usepackage{subcaption}
\usepackage{caption}
\usepackage{xcolor}

\newcommand{\bs}[1]{\boldsymbol{#1}}

\newcommand{\dpar}[2]{\frac{\partial #1}{\partial #2}}

\title{Graph neural networks informed locally by thermodynamics}

\renewcommand{\shorttitle}{Graph neural networks informed locally by thermodynamics}

\hypersetup{
pdftitle={Graph neural networks informed locally by thermodynamics},
pdfsubject={q-bio.NC, q-bio.QM},
pdfauthor={E. Cueto},
pdfkeywords={First keyword, Second keyword, More},
}
\author[1]{Alicia Tierz}
\author[1]{Iciar Alfaro}
\author[1]{David Gonz\'alez}
\author[2,3]{Francisco Chinesta}
\author[1]{El\'ias Cueto}
\affil[1]{{\small ESI Group-UZ Chair of the National Strategy on Artificial Intelligence. \protect\\ Aragon Institute of Engineering Research (I3A). Universidad de Zaragoza. Zaragoza, Spain.}}
\affil[2]{{\small CNRS@CREATE LTD. Singapore.}}
\affil[3]{{\small ESI Group chair. PIMM Lab. ENSAM Institute of Technology. Paris, France. }}

\begin{document}
\maketitle

\begin{abstract}
    Thermodynamics-informed neural networks employ inductive biases for the enforcement of the first and second principles of thermodynamics. To construct these biases, a metriplectic evolution of the physical system under study is assumed. This provides excellent results, when compared to uninformed, black box networks. While the degree of accuracy can be increased in one or two orders of magnitude, in the case of graph networks, this requires assembling global Poisson and dissipation matrices, which breaks the local structure of such networks. In order to avoid this drawback, a local version of the metriplectic biases has been developed in this work, which avoids the aforementioned matrix assembly, thus preserving the node-by-node structure of the graph networks. We apply this framework for examples in the fields of solid and fluid mechanics.  Our approach demonstrates significant computational efficiency and strong generalization capabilities, accurately making inferences on examples significantly different from those encountered during training.
\end{abstract}

\section{Introduction}

Computational simulation is a discipline that has been around for 80 years or so and has emerged as a cornerstone tool across various scientific disciplines, facilitating the prediction of physical phenomena and enabling engineers to refine designs before costly experimental setups are pursued. Traditionally, these simulations have relied heavily on mathematical formulations, often expressed through partial differential equations (PDEs), to model complex systems in fields such as structural mechanics or fluid dynamics \cite{raissi2019physics}. However, with the advent of the information era—the so-called fourth paradigm of science \cite{hey2009fourth}—, a shift towards data-driven approaches, particularly deep learning algorithms, has garnered attention due to their ability to address the limitations of traditional methods, including handling non-linear dynamics under real-time restrictions \cite{moya2022physics}.

Deep learning algorithms, while powerful, are often computationally demanding and require extensive datasets, posing challenges in terms of scalability and generalization \cite{Geiger_2020}. To address these challenges, recent research has explored novel architectures, such as geometric deep learning, which leverage problem structures to enhance performance and reduce data consumption \cite{hamilton2018representation, Bronstein_2017, battaglia2018relational, lee2018attention}. At the same time, traditional mesh-based representations have long been favoured in modelling complex physical systems, offering adaptability and accuracy across various domains, from aerodynamics \cite{airfoils} to structural mechanics \cite{Multiphysics}.

Our work introduces a novel approach that bridges the gap between deep learning, mesh-based simulations, and physics-informed inductive biasing. Leveraging the principles of geometric deep learning, we propose a method that encodes simulation states into graphs, enabling computations in mesh space while ensuring adherence to thermodynamic laws through the use of appropriate inductive biases. Specifically, our contributions can be summarized as follows:
\begin{itemize}
    \item We develop a framework that assumes a metriplectic (metric for the dissipative terms and symplectic for the conservative ones) nodal structure into graph-based representations. This allows us to accurately capture dissipative dynamics while ensuring the fulfilment of thermodynamic principles.
    \item The proposed methodology ensures spatial equivariance, allowing for scalable and resolution-independent learning of physical dynamics.
    \item By leveraging a nodal metriplectic structure, our method improves memory efficiency and computational performance compared to existing approaches, that enforce thermodynamic consistency in a global manner.
    \item Our framework overcomes limitations of previous methods by directly learning dynamics from diverse physical systems while maintaining adherence to the GENERIC formalism at all scales \cite{OettingerDynamics}.
\end{itemize}
This approach removes key bottlenecks in mesh-based simulations, particularly those related to memory usage and computational efficiency, offering a significant improvement in the simulation of high-dimensional physics problems.

\section{Related Work}

Recent advances in data-driven approaches for physics simulation---the so-called neural simulators---have garnered significant interest in the scientific community. One major breakthrough has been the integration of physical principles into machine learning frameworks, aiming to ensure compliance with the fundamental laws of physics. Among these methods, Physics-Informed Neural Networks (PINNs) \cite{raissi2019physics} stand out for embedding partial differential equations (PDEs) governing the system into the learning process. PINNs leverage collocation methods, where the loss function includes a residual term that enforces the satisfaction of the PDEs. While effective, PINNs require explicit knowledge of the governing equations, limiting their applicability in scenarios where such equations are unknown or infeasible to define.

To address these limitations, alternative approaches have drawn parallels between learning physical phenomena and modeling dynamical systems \cite{weinan2017proposal}. These methods frame the problem as non-linear regression tasks that infer time evolution laws directly from data. Such approaches enable the incorporation of known physical properties into learning, ensuring adherence to principles like energy conservation. Hamiltonian neural networks \cite{greydanus2019hamiltonian,mattheakis2022hamiltonian,david2023symplectic,galimberti2023hamiltonian} and Lagrangian neural networks \cite{cranmer2020lagrangian,roehrl2020modeling,bhattoo2023learning} exemplify this strategy, leveraging classical mechanics frameworks to enforce conservation laws. However, these methods primarily focus on conservative systems, neglecting the irreversible and dissipative nature of most real-world phenomena. Addressing this limitation requires embedding additional constraints to ensure compliance with both the first and second laws of thermodynamics.

Beyond PINNs and dynamics-inspired methods, recent research has explored improving efficiency and scalability in simulations. Traditional grid-based methods, which rely on convolutional neural networks (CNNs), dominate the field due to their compatibility with structured data \cite{lecun2015deep}. However, their limited adaptability to unstructured geometries restricts their application in fields like structural mechanics \cite{Multiphysics} and aerodynamics \cite{airfoils}. To overcome these challenges, geometric deep learning (GDL) \cite{bronstein2021geometricdeeplearninggrids} has emerged as a robust framework for modeling irregular and unstructured data. By leveraging the inherent symmetries and structures of the data, GDL methods enable computations on graphs and meshes, expanding the applicability of machine learning in physical simulations.

Within GDL, graph neural networks (GNNs) \cite{hamilton2018representation,Bronstein_2017,battaglia2018relational} have shown significant promise for particle-based systems and mesh-based simulations. GNNs approximate differential operators governing the dynamics of physical systems, making them particularly effective for fluid dynamics \cite{Moya_2022, Ladicky_2015} and granular materials. These approaches have also been combined with classical numerical methods like the Finite Element Method (FEM) \cite{Okumoto2009} to enhance adaptability and accuracy through data-driven corrections. Despite these advancements, many GNN-based methods treat predicted variables as black-box outputs, often lacking explicit physical interpretability.

A complementary direction has focused on integrating physical principles directly into the learning process. The General Equation for the non-Equilibrium Reversible-Irreversible Coupling (GENERIC) formalism \cite{PhysRevE.56.6620} provides a metriplectic framework that combines symplectic and metric structures to describe energy and entropy dynamics. Recent works \cite{Gonzalez_2019, Ghnatios_2019, Moya_2019} have demonstrated the effectiveness of GENERIC in modeling the coarse-grained evolution of complex systems. These developments have led to the introduction of Structure-Preserving Neural Networks (SPNNs) \cite{Hernandez_2021}, which embed GENERIC principles into deep learning models to ensure thermodynamic consistency. SPNNs integrate energy and entropy constraints into the learning process, enabling accurate simulations of dissipative and irreversible phenomena.

Building on these foundations, several recent works have developed methods that embed thermodynamic principles directly into machine learning frameworks. For instance, \cite{Hern_ndez_2021} introduces an approach that combines deep learning with mesh-based simulations and physics-informed inductive biases to ensure compliance with thermodynamic laws. Similarly, \cite{lee2021machine} and \cite{zhang2022gfinns} extend these ideas by leveraging geometric deep learning principles to encode simulation states into graphs. These methods enable computations within mesh-based domains while maintaining spatial equivariance and scalability across resolutions. This framework facilitates the direct learning of dynamics from data, bridging the gap between classical simulation techniques and modern machine learning approaches.

In parallel, machine learning techniques have also shown potential in computational fluid dynamics (CFD). Deep learning accelerates high-dimensional simulations, reducing computational turnaround times and enabling real-time applications in engineering and visualization tasks \cite{inproceedings, Bhatnagar_2019, zhang2018application, Kanwar_2020}. Particle-based representations modeled through deep learning algorithms have proven effective for free-surface liquids and granular materials \cite{Scarselli_GNN, Moya_2022}. These advancements highlight the versatility of machine learning in capturing complex physical behaviors across domains.

Building on these insights, our work integrates graph-based geometric deep learning with the GENERIC formalism to develop a novel framework for simulating physical systems. By embedding thermodynamic inductive biases into GNN architectures, we ensure compliance with energy and entropy evolution while maintaining scalability and efficiency. This approach addresses the limitations of previous methods, offering a unified framework for learning dynamics across diverse physical domains.

\section{Methodology}
\label{sec:methodology}

\subsection{Problem Statement}

Let us assume that the physical phenomenon whose evolution we want to predict is governed by a set of independent state variables $\bs z\in \mathcal{M} \subseteq \mathbb{R}^{D} $, with $\mathcal{M}$ the state space of these variables,
which is assumed to have the structure of a differentiable manifold. These variables encapsulate the essence of the system's dynamics, whose time evolution is given by:
\begin{equation}\label{dyn_sys}
\dot{\bs{z}} =\frac{d\bs z}{dt}= \bs{F}(\bs{z}, t), \quad t \in \mathcal{I} = (0, T], \quad \bs{z}(0) = \bs{z}_0,
\end{equation}
where $t$ refers to the time coordinate in the time interval $\mathcal{I}$ and $\bs F (\bs{z}, t)$ refers to an unknown function governing the system dynamics,  whose precise form is to be unveiled from data.

The primary aim of the study is to identify a suitable mapping function \( \bs{F} \) that can accurately predict the system's behaviour within a predefined time horizon \( T \) using deep learning methodologies. In the absence of any other knowledge on the physical phenomenon---such as its governing partial differential equation--- it is essential that the derived flow map given by $\bs F$ complies, at least, with fundamental thermodynamic principles, ensuring energy conservation and adhering to entropy inequality constraints. This constitutes the so-called ``dynamical system equivalence'' approach to learning physical phenomena from data \cite{weinan2017proposal}.

The proposed solution aims not only to accurately and efficiently predict the time evolution of simple and complex systems but also to provide information about the underlying physics governing the system dynamics. By meeting basic thermodynamic requirements and taking advantage of advanced machine learning techniques, our approach aims to improve the understanding and predictive capability of complex dynamical systems. 

To this end we propose, as in previous works \cite{hernandez2022thermodynamics}, using both geometric and thermodynamic inductive biases to improve the accuracy and generalization of the resulting integration scheme. The first is achieved with Graph Neural Networks (GNNs), which induce a non-Euclidean geometric prior with permutation invariant node and edge update functions. The second bias is enforced by learning an assumed GENERIC structure of the problem to model general non-conservative dynamics.

\subsection{Geometric structure: Graph Neural Networks}

\begin{figure}[h]
    \centering
    \includegraphics[scale=0.5]{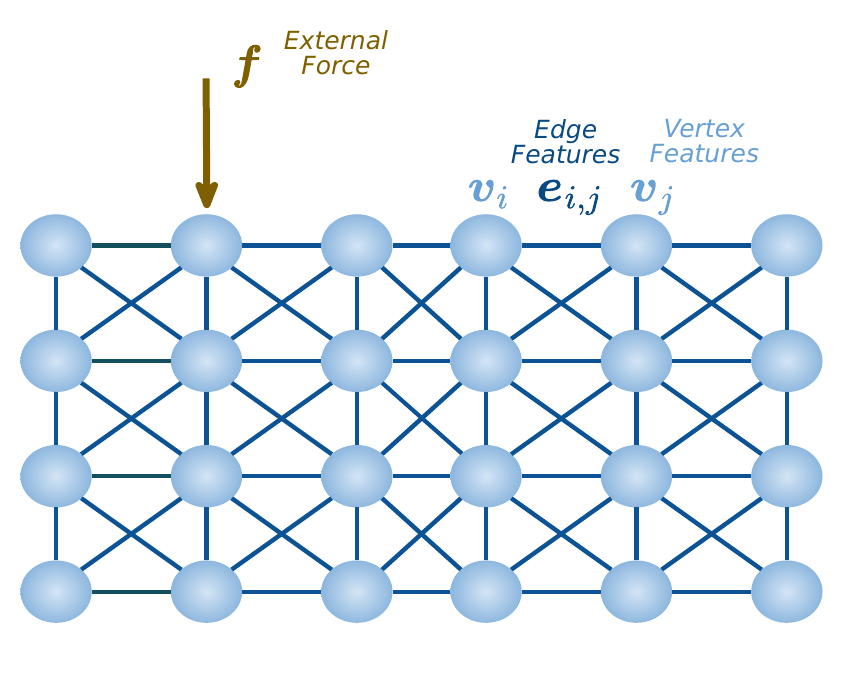}
    \caption{Scheme of the discretisation of a physical system by converting a mesh into a graph with which to train our GNN.}
    \label{fig:graph_scheme}
\end{figure}

As we have just explained in the previous section, the geometric bias is ensured through the use of graph neural networks (GNNs). We construct a directed graph $\mathcal{G} = (\mathcal{V},\mathcal{E}, \bs u)$, where  $\mathcal{E}\subseteq \mathcal{V} \times  \mathcal{V}$ is a set of  $\left| \mathcal{E} \right| = e$ edges and $\mathcal{V} = \left\{ 1, ..., n \right\}$ is a set of $\left| \mathcal{V} \right| = n$ vertices. Each vertex and edge in the graph represents a node and a pairwise interaction between nodes in a simplified physical system. For each vertex $i\in\mathcal{V}$, we have a feature vector $\bs v_{i}\in\mathbb{R}^{F_{g}}$ representing the physical properties of that node. Similarly, for each edge $(i, j)\in\mathcal{E}$, there's an edge feature vector $\bs e_{ij}\in\mathbb{R}^{F_{g}}$. When working with synthetic data coming from finite element, finite volume or finite differences, there is a straightforward one-to-one equivalence between the mesh structure of these methods and the graphs in GNNs. We assume a one-to-one correspondence between finite element (also finite difference or related methods) and the vertices of the graph. However, we do not conserve finite element connectivity: we create instead edges that connect pairs of nodes based on a specific distance threshold. This is motivated by the message passing algorithm that will be described below, see Fig. \ref{fig:graph_scheme}.

In our setup, we assign the positional state variables of the system $(\bs q_{i})$ to the edge feature vector $\bs e_{ij}$. This means that the edge features represent relative distances $(\bs q_{ij} = \bs q_{i} - \bs q_{j})$ between nodes, giving the graph network a focus on distance-based interactions and keeping it invariant under translation and rotation.
Other state variables are assigned to the node feature vector $\bs v_{i}$. External interactions, like forces acting on the system, are included in an external load vector $\textbf{f}_{i}$.

We illustrate this graph representation of a physical system in Figure  \ref{fig:graph_scheme}, showing node state variables, relative nodal distances and external interactions.

These features are then processed using an encode-process-decode scheme, involving several multilayer perceptrons (MLPs) shared among all nodes and edges of the graph. This algorithm consists of five steps and is described in more detail in Section \ref{subsec:Learning}.

\subsection{Metriplectic structure: The GENERIC formalism. }

The implementation of thermodynamic inductive biases relies on the General Equation for Non-Equilibrium Reversible-Irreversible Coupling (GENERIC) formalism \cite{PhysRevE.56.6620, OettingerDynamics}. This framework extends the classical Hamiltonian formulation to dissipative systems, providing a generalized approach to modeling general system dynamics. Within this framework, reversible or conservative contributions follow Hamiltonian principles, requiring an energy potential and a Poisson bracket structure.

Under the GENERIC formulation of time evolution for non-equilibrium systems, characterized by the set of state variables $\bs z$, the evolution of the system is no longer the unknown function $\bs F$ of Eq. (\ref{dyn_sys}), but is assumed to be of the form:
\begin{equation} \label{eq:bracket_form}
\frac{d\bs z}{dt} = \{ \bs z, E \} + \left[ \bs z, S \right],
\end{equation}
where $\{ \cdot,\cdot\}$ denotes the classic Poisson bracket and $[\cdot,\cdot]$ represents the dissipative bracket. To facilitate practical use, these brackets are often reformulated using two linear operators:
\begin{equation} \label{eq:reformulation}
\mbox{\boldmath$L$}:T^{*}\mathcal{M} \rightarrow T\mathcal{M}, \quad \mbox{\boldmath$M$}:T^{*}\mathcal{M} \rightarrow T\mathcal{M},
\end{equation}
where $T^{*}\mathcal{M}$ and $T\mathcal{M}$ denote the cotangent and tangent bundles of the state space $\mathcal{M}$, respectively. The operator $\bs L = \bs L (\bs z)$ corresponds to the Poisson bracket and must be skew-symmetric. Similarly, the friction matrix $\bs M = \bs M (\bs z)$ captures the irreversible behavior of the system and is symmetric and positive semi-definite, ensuring a non-negative dissipation rate.

Given these components and considering a system discretized into $\tt n_v$ particles, the system must adhere to the laws of thermodynamics. To enforce this, we assume a metriplectic evolution, whose precise form is to be found from data:
\begin{equation}\label{GENERIC2}
\dot{ \bs z} = \bs L( \bs z) \frac{\partial E}{\partial \bs z} + \bs M( \bs z)\frac{\partial S}{\partial \bs z},
\end{equation}
where $\bs z \in \mathbb{R}^{\tt n_v \times \tt n_{\text{dof}}}$ and $\tt n_{\text{dof}}$ represents the number of degrees of freedom of each particle.

To ensure the fulfillment of the first and second principles of thermodynamics, we must also satisfy the so-called degeneracy conditions,
\begin{equation}\label{deg1}
\bs L(\bs z)\frac{\partial S}{\partial \bs z} = \bs 0,
\end{equation}
and
\begin{equation}\label{deg2}
\bs M(\bs z)\frac{\partial E}{\partial \bs z} = \bs 0.
\end{equation}
This is so, since
\begin{equation}\label{eq:Econs}
 \dpar{E}{t} = \dpar{E}{\bs{z}}\cdot\dpar{\bs{z}}{t}=\dpar{E}{\bs{z}}\left( \bs{L} \dpar{E}{\bs{z}} + \bs{M} \dpar{S}{\bs{z}}\right)=0,
\end{equation}
which expresses the conservation of energy in an isolated system---in other words, the first law of thermodynamics---. The same holds for the entropy $S$, which in turn satisfies:
\begin{equation}\label{eq:Scons}
 \dpar{S}{t} = \dpar{S}{\bs{z}}\cdot\dpar{\bs{z}}{t}=\dpar{S}{\bs{z}}\left( \bs{L} \dpar{E}{\bs{z}} + \bs{M} \dpar{S}{\bs{z}}\right)=\dpar{S}{\bs{z}}\bs{M}\dpar{S}{\bs{z}}\geq 0,
\end{equation}
that is, the second law of thermodynamics.

Given $N_{\text{sim}}$ experimental data sets $\mathcal{D}_{\text{sim}}$ containing labelled pairs of a single-step state vector $\bs{z}(t)$ and its evolution in time $\bs{z}({t+\Delta t})$,
\begin{equation}
\mathcal{D}=\{\mathcal{D}_{\text{sim}}\}_{{\text{sim}}=1}^{N_{\text{sim}}},\quad\mathcal{D}_{\text{sim}} =\{(\bs{z}(t),\bs{z}({t+\Delta t}))\}_{t=0}^{T},
\end{equation}
we construct a neural network by considering two different loss terms. First, a data-loss term that takes into account the correctness of the network prediction of the state vector at subsequent time steps by integrating GENERIC in time, i.e.,
\begin{equation*} \label{ec:lossdata}
\mathcal{L}^{\text{data}}=\left\Vert\dot{\bs{z}}^{\text{GT}}-\dot{\bs{z}}^{\text{net}}\right\Vert^2_2,
\end{equation*}
with $\Vert\cdot\Vert_2$ the L2-norm, $\dot{\bs{z}}^{\text{GT}}$ the ground truth solution which is computed using the same time integration scheme employed to discretize Eq. (\ref{eq:bracket_form}) and $\dot{\bs{z}}^{\text{net}}$ the network prediction. The choice of the time derivative instead of the state vector itself is employed to regularize the global loss function to a uniform order of magnitude with respect to the degeneracy terms.

We then consider a second loss term to take into account the fulfilment of the degeneracy equations in a soft way,
\begin{equation*} \label{ec:lossdeg}
\mathcal{L}^{\text{deg}}=\left\Vert\bs{L}\dpar{S}{\bs{z}}\right\Vert^2_2+\left\Vert\bs{M}\dpar{E}{\bs{z}}\right\Vert^2_2.
\end{equation*}

The total loss term is calculated as the sum of these two terms, with a hyperparameter $\lambda$ that balances the contribution of each one:
\begin{equation}  \label{ec:loss}
\mathcal{L}= \mathcal{L}^{\text{deg}} + \lambda \mathcal{L}^{\text{data}}.
\end{equation}

\subsection{Limitations of thermodynamics-informed graph neural networks}

Graph neural networks are well known for their good properties but also for their high computational cost and poor scalability \cite{ma2022graph}.

The imposition of the metriplectic formalism as an inductive bias in a neural network setting additionally induces the need to assemble  (and store in memory) two matrices of size $(\tt n_v \times \tt n_{\text{dof}}) \times (\tt n_v \times \tt n_{\text{dof}})$. For a system with a large number of particles ($ \tt n_v$ does not exceeed $\mathcal O(10^3)$ particles, with $n_{\text{dof}}$ betwwen 7 to 12  in the presented examples), this leads to big matrices that partially or totally undermine the advantages of the local, vertex-based architecture of graph neural networks. This is of course far from today's standard in finite element computations, where models of tens of millions of degrees of freedom are not infrequent.

Employing sparse matrices to store $\bs L$ and $\bs M$ does not alleviate these limitations, since most deep learning libraries reconstruct the full matrix structure to execute the backpropagation algorithms. The result is a high memory demand and a slowdown of the process, if not a failure in the process due to lack of memory. 

Beyond this difficulty, the structure of the metriplectic formalism destroys the local character of the architecture of graph neural networks. This is precisely one of their most interesting ingredients for the type of applications we pursue.

To address these memory and processing issues, in the next section we describe a local, nodal implementation of the metriplectic bias that conserves the good properties of the graph architecture. This new approach provides more efficient solutions, avoiding the need for storing large matrices and enabling smoother training of graph neural networks in thermodynamics-informed applications.

\section{A local implementation of thermodynamics-informed GNNs}
\label{sec:local_gen}

\subsection{Port-metriplectic formalism}

The previous formulation, Eqs. (\ref{GENERIC2}) and (\ref{deg1})-(\ref{deg2}), assume inherently that the system is closed, i.e., that there is no energy exchange with the environment. For open systems, an alternative formulation should be employed \cite{hernandez2023port}.

If we try to keep the nodal structure of graph neural networks, it is obvious that each vertex of the graph is an open system, as it exchanges energy with neighbouring nodes. For this reason, in this paper we have developed a local formulation of the inductive bias that follows a port-metriplectic formalism. In it, each node will represent a portion of the bulk, while exchanging energy with the neighbouring nodes through predefined ports (i.e., the edges of the graph). In this setting, the Poisson and dissipative brackets take the form
\begin{equation}
    \lbrace\cdot,\cdot\rbrace = \lbrace\cdot,\cdot\rbrace_{\text{bulk}} + \lbrace\cdot,\cdot\rbrace_{\text{boun}},
\end{equation}
and
\begin{equation}
    [\cdot,\cdot] = [\cdot,\cdot]_{\text{bulk}} + [\cdot,\cdot]_{\text{boun}}.
\end{equation}
In other words, both brackets are decomposed additively into bulk and boundary contributions. With this decomposition in mind, the GENERIC principle, Eq. (\ref{GENERIC2}), applies to the bulk portion of the system associated to each node, that now reads
\begin{equation}\label{PM}
    \dot{\bs z} = \lbrace \bs z,E\rbrace_{\text{bulk}} + [\bs z,S]_{\text{bulk}} 
    = \lbrace \bs z,E\rbrace + [\bs z,S] - \lbrace \bs z,E\rbrace_{\text{boun}} - [\bs z,S]_{\text{boun}}.
\end{equation}

In matrix form, we obtain
\begin{equation}\nonumber
\dot{\bs z}  = \bs L \frac{\partial E}{\partial \bs z} + \bs M\frac{\partial S}{\partial \bs z}  - \bs L_{\text{boun}} \frac{\partial E_{\text{boun}}}{\partial \bs z} - \bs M_{\text{boun}}\frac{\partial S_{\text{boun}}}{\partial \bs z}.\label{eq:port}
\end{equation}
In this case, the degeneracy conditions hold at the bulk level only,
\begin{equation}\label{deg12}
\bs L_{\text{bulk}}(\bs z)\frac{\partial S_{\text{bulk}}}{\partial \bs z} = \bs 0,
\end{equation}
and
\begin{equation}\label{deg22}
\bs M_{\text{bulk}}(\bs z)\frac{\partial E_{\text{bulk}}}{\partial \bs z} = \bs 0.
\end{equation}

So if we plan to impose a GENERIC-like structure at a particle level, we must assume that this particle receives energy input/outputs from surrounding particles (as dictated by the graph structure). 

Assume that a given particle $i$ with state variables $\bs z_i \in \mathbb R^{\tt n_{\text{dof}}}$ is connected in the graph with $j=1, \ldots, \tt n_{\text{neigh}}$ neighbouring particles. Its state variables will thus evolve in time as
\begin{equation}
\dot{\bs z}_i  = \bs L_i (\bs z_i)  \frac{\partial e_i}{\partial \bs z_i} + \bs M_i(\bs z_i)\frac{\partial s_i}{\partial \bs z_i}  - \sum_j^{\tt n_{\text{neigh}}} \left[ \bs L_{ij} (\bs z_j)\frac{\partial e_j}{\partial \bs z_j} + \bs M_{ij}(\bs z_j) \frac{\partial s_{j}}{\partial \bs z_j} \right].\label{eq:port2}
\end{equation}

The total energy and entropy of the system are obtained by assuming
$$
E = \sum_i^{\tt n_v} e_i,
$$
and 
$$
S=\sum_i ^{\tt n_v} s_i.
$$

The degeneracy conditions at the bulk level play here the role of the particle level, so that
\begin{equation}\label{deg13}
\bs L_{i}(\bs z_i)\frac{\partial s_{i}}{\partial \bs z_i} = \bs 0,
\end{equation}
and
\begin{equation}\label{deg23}
\bs M_{i}(\bs z_i)\frac{\partial e_{i}}{\partial \bs z_i} = \bs 0,
\end{equation}
for all $i=1, \ldots, \tt n_v$.

We could envisage additional restrictions, such as total energy $E$ conservation at the system level or total entropy $S$ production to be non-negative, for instance. These restrictions, in principle, can be easily imposed through the loss function, in soft way. However, our experiments show excellent results without the need to impose them. As in many other scenarios, there is a balance between physical restrictions and expressivity of the networks that should be taken into account and for which there is no simple (nor single) recipe.

\subsection{Learning procedure}
\label{subsec:Learning}

\begin{figure}[h]
    \centering
    \includegraphics[width=1\textwidth]{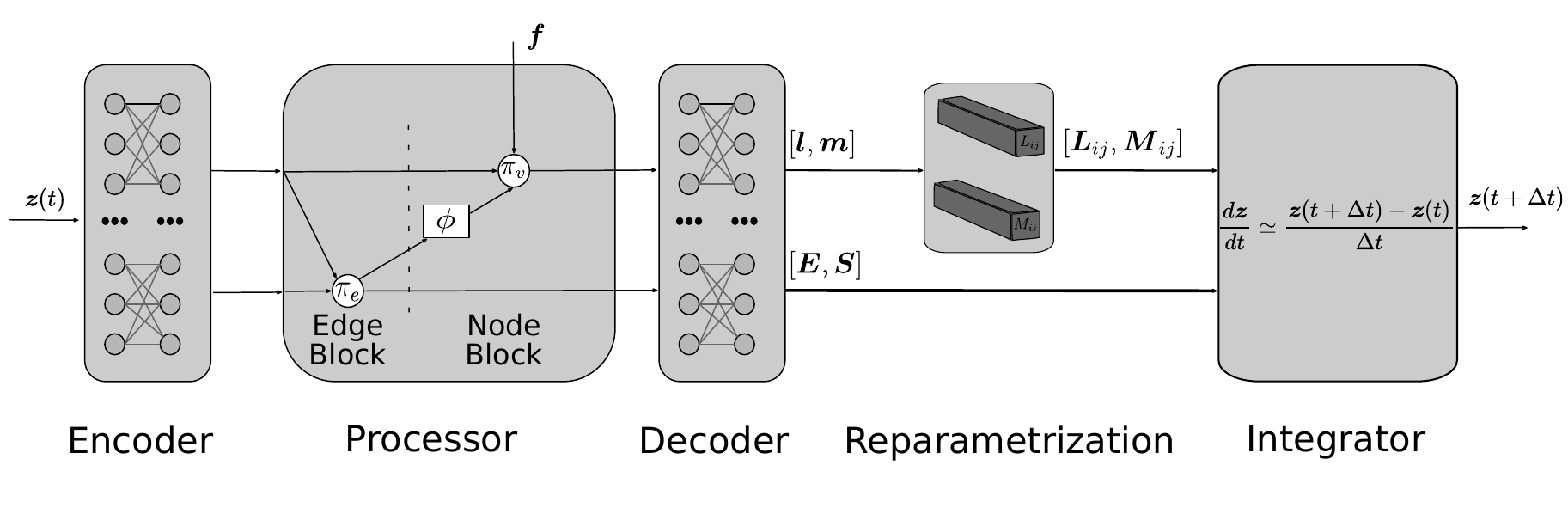}
    \caption{Scheme of the information processing within a thermodynamics-informed Graph Neural Network.}
    \label{fig:spnn_scheme}
\end{figure}

This section aims to consolidate all the progress made thus far. The algorithm we will utilize comprises five steps, illustrated in Figure \ref{fig:spnn_scheme}.

The first step involves the \textbf{encoder}, which consists of two multi-layer perceptrons (MLPs) tasked with transforming the feature vectors of both nodes and edges into a higher-dimensional space. This transformation converts a mesh composed of nodes and edges into a bidirectional graph. For the edge feature vector, we incorporate the relative displacement in mesh space as $\bs u_{ij} = \bs u_i - \bs u_j$ and its magnitude $|\bs u_{ij}|$ to achieve Galilean invariance. Additionally, the node feature vector covers the remaining physical characteristics of the nodes, such as energy, stress, or velocity.

Once we have the feature vectors in latent space, we proceed to the \textbf{processor} component. This phase of the algorithm involves the exchange of nodal and edge information among vertices and connections through a process known as message passing. During this step, each vertex aggregates information from its neighboring vertices and updates its own feature vector accordingly. Similarly, each edge relays information between its connected vertices, facilitating the exchange of relevant data across the graph structure. Additionally, in the nodal component, external forces applied to the system are taken into account, ensuring that the model captures the impact of external influences on the dynamics of the system. This iterative message passing process allows for the refinement and integration of information across the entire graph, enabling the network to capture complex relationships and dependencies within the data.

At the output of the processor, we have the \textbf{decoder} component, which comprises four small decoders, each implemented with an MLP that transforms the latent node and edge features into output features. In this approach, we predict the GENERIC energy $\frac{\partial e_i}{\partial \bs z_i}$ and entropy $\frac{\partial s_i}{\partial \bs z_i}$ gradients of each particle from the output of the node block. With the other two decoders, we predict the $\bs l$ and $\bs m$ flattened operators of each edge. For these latter two decoders, the input includes both nodal and edge features.

Once we have the flattened $\bs m_{i}$, $\bs l_{i}$, $\bs m_{ij}$  and $\bs l_{ij}$ operators, we need to reshape them into lower-triangular matrices to obtain $\bs M_{i}$, $\bs L_{i}$, $\bs M_{ij}$ and $\bs L_{ij}$ in the \textbf{reparametrization} part. The positive semi-definite and skew symmetric conditions are imposed during construction using the following parameterization:
\begin{equation}\label{conf_lm}
\bs M_{ij}=\bs m_{ij}\bs m_{ij}^{\top},   \qquad   \bs L_{ij}= \bs l_{ij}- \bs l_{ij}^{\top}.
\end{equation}

In the final step, we utilize the metriplectic structure as a soft constraint to construct a thermodynamically-sound time \textbf{integrator}, which serves as the thermodynamic inductive bias of our approach. This single-step integration is carried out using a forward Euler scheme with a time increment $\Delta t$, incorporating the nodal GENERIC formalism, resulting in 
\begin{equation}\label{intgration}
\bs z_i (t+\Delta t) = \bs z_i (t) + \dot{ \bs z_i}(t) \Delta t  .
\end{equation}

Other aspects that also impact the learning process, yet fall outside the algorithm itself, include the choice of activation functions, optimization methods, and the addition of noise to augment the datasets.

As activation functions, we employ the Swish function \cite{ramachandran2017searching} in our approach. This function integrates non-zero second derivatives with ReLU-type nonlinearities, aligning with our need to predict $\frac{\partial E}{\partial \bs z}$ and $\frac{\partial S}{\partial \bs z}$. This selection ensures proper backpropagation of weights and biases, crucial for maintaining continuity in the activation functions.

During optimization, all training processes utilize the Adam optimizer \cite{kingma2014adam}, complemented by a multistep learning rate scheduler that progressively decreases the learning rate throughout training periods.

To enhance learning, the network inputs are normalized to a range between zero and one. Each state variable is normalized separately to preserve their individual scales and dynamics, ensuring that the network can effectively process and learn from the distinct characteristics of each variable. Furthermore, Gaussian noise is applied after this normalization during training to simulate error accumulation during time integration \cite{pfaff2021learning}. Another factor that significantly impacted training results was the utilization of low batch size, as discussed in the aforementioned reference.

The trained networks are implemented in PyTorch and are publicly available online, along with the datasets, at \href{https://github.com/a-tierz/Local-ThermodynamicsGNN}{{\color{blue}https://github.com/a-tierz/Local-ThermodynamicsGNN}}.

\section{Numerical experiments}
\label{sec:exps}
We evaluated our method on different examples involving solid and fluid mechanics. Training and test synthetic data were obtained through high-fidelity finite element simulations. The resulting dataset $\mathcal D$ was partitioned into $80 \% $ training, $10 \% $ test, and $10 \% $ validation sets.
The proposed experiments will evaluate and compare three different models: the SPNN (Structure-Preserving Neural Network) \cite{Hern_ndez_2021}, which consists of an encoder, decoder, and a processor where the GENERIC formalism is embedded; a vanilla GNN which serves as a baseline without any thermodynamic bias; and our proposed method the nodal TIGNN (Thermodynamic Informed Graph Neural Network), which incorporates both geometric and thermodynamic biases. Additionally, in the first experiment, we include the results of the global TIGNN implementation for comparison.

 To assess the model's performance, we measure the mean error during rollouts for the different state variables for all nodes. The metrics used are the Root Mean Squared Error, 
\begin{equation}
\text{RMSE} = \sqrt{\frac{1}{n}\sum_{i=1}^{n} \| \bs y_{i} - \hat{\bs y}_{i} \|^2},
\label{eq:rmse}
\end{equation}
 and the Relative Root Mean Squared Error, as given by 
\begin{equation}
\text{RRMSE} = \sqrt{\frac{1}{n} \sum_{i=1}^{n} \left( \frac{\| \bs y_i - \hat{\bs y}_i \|}{\| \hat{\bs y}_i \|_{\infty}} \right)^2},
\label{eq:rrmse_inf}
\end{equation}
where $\bs y$ represents the output variable.
The error bars denote the standard error across different trajectories.

Documentary videos of the obtained rollouts can also be found in the mentioned GitHub repository.

\subsection{3D Viscoelastic Beam Bending}
\label{subsec: 3DBeam}
For this first experiment, we utilized a publicly available dataset obtained by Q. Hernandez and coworkers \cite{hernandez2022thermodynamics}. The dataset consists of 52 simulations of a viscoelastic cantilever beam subjected to bending forces. The material properties are characterized by a single-term polynomial strain energy potential, represented by the equation:
\begin{equation*}
U = C_{10}(\bar{I}_{1}-3)+C_{01}(\bar{I}_{2} -3)+\frac{1}{D_{1}}(J_{el}-2)^{2},
\end{equation*}
where $U$ is the strain potential, $C_{10}$ and $C_{01}$ are shear material constants, $I_{1}$ and $I_{2}$ are the two invariants of the left Cauchy-Green deformation tensor and $D_{1}$ is the material compressibility parameter.  $J_{el}$ represents the elastic volume ratio. The viscoelastic component is described by a two-term Prony series of the dimensionless shear relaxation modulus,
\begin{equation*}
g_{R}(t) = 1-\bar{g}_1(1-e^{-\frac{t}{\tau_{1}}})-\bar{g}_2(1-e^{-\frac{t}{\tau_{2}}}),
\end{equation*}
with relaxation coefficients of $\bar{g}_1$ and $\bar{g}_1$, and relaxation times of $\tau_{1}$ and $\tau_{2}$.

The state variables for each node of the viscoelastic beam include position $\bs q$, velocity $\bs v$, and stress tensor $\bs \sigma$:
\begin{equation*}
    \mathcal S =\left\{ \bs z=(\bs q,\bs v,\bs \sigma)\in \mathbb{R}^{3}\times \mathbb{R}^{3} \times \mathbb{R}^{6} \right\}.
\end{equation*}

The edge feature vector incorporates the relative deformed position, while the remaining variables are included in the node feature vector. To differentiate between encastred and free nodes, we augmented the node features with an additional one-hot vector, denoted as $\bs n$. Furthermore, the external load vector $\bs f$ was incorporated into the node processor MLP as an external interaction.

\begin{figure}[h]
    \centering
    \includegraphics[width=\textwidth]{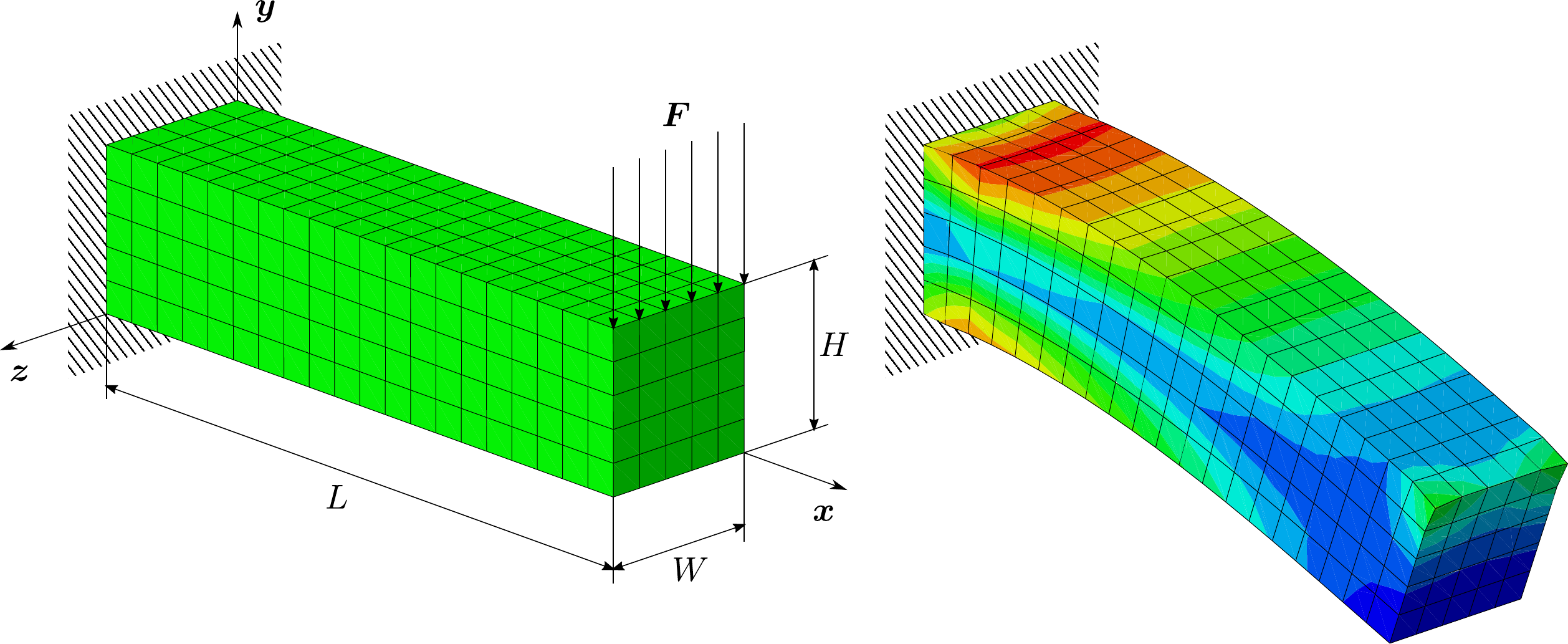}
    \caption{Sketch of the viscous-hyperelastic 3D beam bending problem.}
    \label{beam}
\end{figure}

The dataset is composed of $N_{\text{sim}} = 52$ simulations of  3D bending beams. The prismatic beam dimensions are $H = 10$, $W = 10$, and $L = 40$, discretized into $N = 756$ nodes. The material's hyperelastic and viscoelastic parameters are $C_{10} = 1.5\cdot 10^5$, $C_{01} = 5 \cdot 10^3$, $D_1 = 10^{-7}$, $\bar{g}_{1} = 0.3$, $\bar{g}_{2} = 0.49$, $\tau_{1} = 0.2$, $\tau_{2} = 0.5$, respectively. A distributed load of $F = 10^5$ is applied at different positions with an orientation perpendicular to the solid surface, see Fig. \ref{beam}. The 52 simulations are discretized into $N_T = 20$ time increments of $\Delta t = 5 \cdot 10^{-2}$.

The chosen hyperparameters for training include a hidden dimension of $F_h = 80$ for the node and edge latent vectors, 9 passes through the processor part, and a batch size of 4. The learning rate is set to $lr = 8 \times 10^{-4}$ with decreasing order of magnitude 0.3 at epochs 100, 250, and 500, over a total of $N_{\text{epoch}} = 1000$ epochs. Additionally, the training noise variance is set to $\sigma_{\text{noise}} = 4 \times 10^{-5}$.

\begin{figure}[h]
    \centering
    \includegraphics[width=\linewidth]{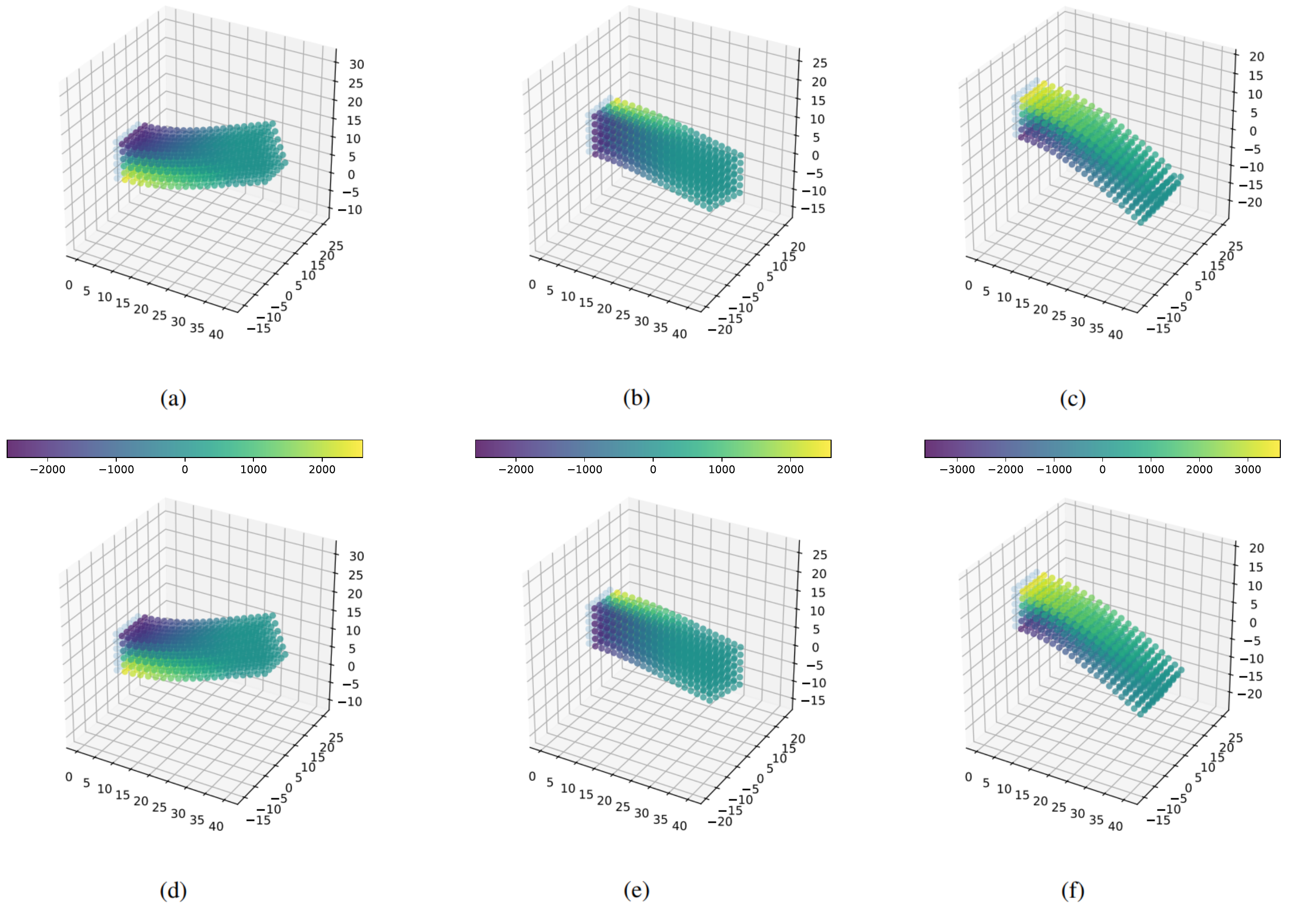}
    \caption{The results depict the predictions for the three test simulations. Figures (a), (b), and (c) illustrate the network's inferred outputs at sample time instants, while figures (d), (e), and (f) depict the ground truth for comparison. The color coding denotes the $xx$ component of the stress tensor. }
    \label{fig:beam3D_rollout}
\end{figure}

The results from the test data are illustrated in Fig. \ref{fig:beam3D_resutls}. The boxplots depict the outcomes of four models: the blue boxplots correspond to the SPNN model, the light grey correspond to the vanilla GNN architecture, the burgundy boxplots show the results of global TIGNN, and the yellow boxplots show the results of our nodal port-metriplectic bias approach. The incorporation of the thermodynamic inductive bias yields a significant improvement in predictive performance, as expected. The obtained accuracy improvement is entirely similar to that of the global implementation, see \cite{hernandez2022thermodynamics}. Additionally, Fig. \ref{fig:beam3D_rollout} presents three rollouts of test simulations, where the ground truth is compared with our network's predictions.

\begin{figure}[]
    \centering 
    \begin{tikzpicture}
        \begin{axis}[
            boxplot/draw direction=y,
            ymajorgrids=true,
            height=8cm, 
            width=8cm,  
            axis x line* = bottom,
            axis y line* = left,
            ylabel={Root Mean Squared Error},
            ymode=log, 
            cycle list={{teal_blue, fill=teal_blue, fill opacity=0.3},
                        {ash_gray, fill=ash_gray, fill opacity=0.3},
                        {burgundy, fill=burgundy, fill opacity=0.3},
                        {mustard, fill=mustard, fill opacity=0.3}}, 
            boxplot={
                draw position={1/4 + 1*floor(\plotnumofactualtype/4) + 1/4*mod(\plotnumofactualtype,4)},
                box extend=0.15 
            },      
            xtick={0.5,1.5,2.5},
            xticklabels={%
                {$\bs q$},%
                {$\dot{\bs q}$},%
                {$\bs \sigma$},%
            },
            x tick label style={
                text width=2.5cm,
                align=center
            },
        ]
        \addplot table[row sep=newline, col sep=tab, y index=0] {PLANTILLA_INTERNET/data/mse_output_beam3D.txt};
        \addplot table[row sep=newline, col sep=tab, y index=1] {PLANTILLA_INTERNET/data/mse_output_beam3D.txt};
        \addplot table[row sep=newline, col sep=tab, y index=2] {PLANTILLA_INTERNET/data/mse_output_beam3D.txt};
        
        \addplot table[row sep=newline, col sep=tab, y index=3] {PLANTILLA_INTERNET/data/mse_output_beam3D.txt};
        \addplot table[row sep=newline, col sep=tab, y index=4] {PLANTILLA_INTERNET/data/mse_output_beam3D.txt};
        \addplot table[row sep=newline, col sep=tab, y index=5] {PLANTILLA_INTERNET/data/mse_output_beam3D.txt};

        \addplot table[row sep=newline, col sep=tab, y index=6] {PLANTILLA_INTERNET/data/mse_output_beam3D.txt};
        \addplot table[row sep=newline, col sep=tab, y index=7] {PLANTILLA_INTERNET/data/mse_output_beam3D.txt};
        \addplot table[row sep=newline, col sep=tab, y index=8] {PLANTILLA_INTERNET/data/mse_output_beam3D.txt};

        \addplot table[row sep=newline, col sep=tab, y index=9] {PLANTILLA_INTERNET/data/mse_output_beam3D.txt};
        \addplot table[row sep=newline, col sep=tab, y index=10] {PLANTILLA_INTERNET/data/mse_output_beam3D.txt};
        \addplot table[row sep=newline, col sep=tab, y index=11] {PLANTILLA_INTERNET/data/mse_output_beam3D.txt};
        \end{axis}
        \begin{axis}[
            boxplot/draw direction=y,
            ymajorgrids=true,
            height=8cm, 
            width=7cm,  
            axis x line* = bottom,
		axis y line* = left,
            ylabel={Relative Root Mean Squared Error},
            ymode=log, 
            cycle list={{teal_blue, fill=teal_blue, fill opacity=0.3},
                        {ash_gray, fill=ash_gray, fill opacity=0.3},
                        {burgundy, fill=burgundy, fill opacity=0.3},
                        {mustard, fill=mustard, fill opacity=0.3}}, 
            boxplot={
                draw position={1/4 + 1*floor(\plotnumofactualtype/4) + 1/4*mod(\plotnumofactualtype,4)},
                box extend=0.15 
            },       
            xtick={0.5,1.5,2.5},
            xticklabels={%
                {$\bs q$},%
                {$\dot{\bs q}$},%
                {$\bs \sigma$},%
            },
            x tick label style={
                text width=2.5cm,
                align=center
            },
            at={(8cm,0)}, 
        ]
        \addplot table[row sep=newline, col sep=tab, y index=0] {PLANTILLA_INTERNET/data/mse_inf_output_beam3D.txt};
        \addplot table[row sep=newline, col sep=tab, y index=1] {PLANTILLA_INTERNET/data/mse_inf_output_beam3D.txt};
        \addplot table[row sep=newline, col sep=tab, y index=2] {PLANTILLA_INTERNET/data/mse_inf_output_beam3D.txt};
        
        \addplot table[row sep=newline, col sep=tab, y index=3] {PLANTILLA_INTERNET/data/mse_inf_output_beam3D.txt};
        \addplot table[row sep=newline, col sep=tab, y index=4] {PLANTILLA_INTERNET/data/mse_inf_output_beam3D.txt};
        \addplot table[row sep=newline, col sep=tab, y index=5] {PLANTILLA_INTERNET/data/mse_inf_output_beam3D.txt};
        
        \addplot table[row sep=newline, col sep=tab, y index=6] {PLANTILLA_INTERNET/data/mse_inf_output_beam3D.txt};
        \addplot table[row sep=newline, col sep=tab, y index=7] {PLANTILLA_INTERNET/data/mse_inf_output_beam3D.txt};
        \addplot table[row sep=newline, col sep=tab, y index=8] {PLANTILLA_INTERNET/data/mse_inf_output_beam3D.txt};

        \addplot table[row sep=newline, col sep=tab, y index=9] {PLANTILLA_INTERNET/data/mse_inf_output_beam3D.txt};
        \addplot table[row sep=newline, col sep=tab, y index=10] {PLANTILLA_INTERNET/data/mse_inf_output_beam3D.txt};
        \addplot table[row sep=newline, col sep=tab, y index=11] {PLANTILLA_INTERNET/data/mse_inf_output_beam3D.txt};
        \end{axis}
        \begin{axis}[hide axis,
            xmin=0, xmax=1, ymin=0, ymax=1,
            legend style={
            at={(0.8,-0.2)},  
            anchor=north,     
            legend columns=4, 
            /tikz/every node/.append style={align=center} 
            }           
        ]
            \addlegendimage{area legend, teal_blue, fill=teal_blue, fill opacity=0.2, only marks, mark=square*}
            \addlegendentry{SPNN}
            \addlegendimage{area legend, ash_gray, fill=ash_gray, fill opacity=0.2, only marks, mark=square*}
            \addlegendentry{GNN}
            \addlegendimage{area legend, burgundy, fill=burgundy, fill opacity=0.2, only marks, mark=square*}
            \addlegendentry{global-TIGNN}
            \addlegendimage{area legend, mustard, fill=mustard, fill opacity=0.2, only marks, mark=square*}
            \addlegendentry{nodal-TIGNN}
        \end{axis}
    \end{tikzpicture}
    \caption{
   Zero-shot performance for 4 different trajectories. RMSE and RRMSE values for position ($ \bs q$), velocity $ \dot{\bs q}$ and von Mises stress (\(\bs \sigma\)). Each point on the plot represents the error an individual trajectory. The blue boxes represent the error distribution for the SPNN, the light grey boxes show the distribution for the vanilla-GNNS, the burgundy boxes show the error distribution for the global TIGNN and, finally, the yellow boxes represent the results for the just proposed technique.
    }
    \label{fig:beam3D_resutls}
\end{figure}

Finally, Table \ref{tab:beam3D_computationalcost} presents a comparison of training and inference times for the different models. The results demonstrate that, while the complexity of the nodal-TIGNN architecture leads to increased computational times, the growth in time is significantly smaller compared to a global-TIGNN implementation. This highlights the efficiency of the nodal approach in balancing computational cost with model complexity. For comparison,
running the same simulation in Abaqus requires approximately 22 seconds, which is significantly longer than the inference time of the presented networks.

\begin{table}[h]
    \centering
    \begin{tabular}{|c|c|c|c|c|}
        \hline
        & SPNN& GNN&  Nodal TIGNN &Global TIGNN\\
        \hline
        Training time per epoch (s)& 6& 8& 9.55&91.25\\
        \hline
        Inference time (s)& 8.40& 11.88& 14.628&119.118\\\hline
    \end{tabular}
    \vspace{0.3cm}
    \caption{
      Comparison of computational times for the three models. The table shows the training time required per epoch and the inference time for 20 frames. The results highlight the computational efficiency of the Nodal-TIGNN model compared to the Global-TIGNN implementation, while still outperforming the SPNN and GNN in terms of generalization and accuracy.
    }
    \label{tab:beam3D_computationalcost}
\end{table}

\subsection{2D Viscoelastic bending beam}
\label{subsec: 2DBeam}

\subsubsection{Description of the experiment}

\begin{figure}[h]
    \centering
    \includegraphics[width=\textwidth]{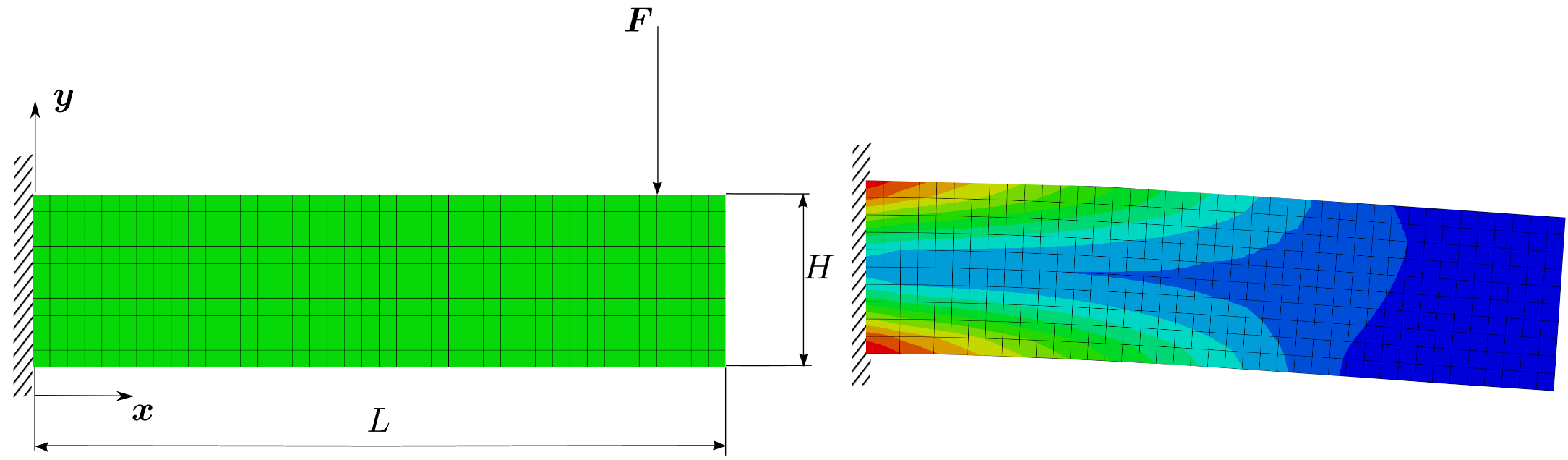}
    \caption{Sketch of the viscous-hyperelastic 2D beam bending problem.}
    \label{beam2D}
\end{figure}

The next experiment also consists of a viscoelastic cantilever beam but this time in 2D and with different geometric properties, see Fig. \ref{beam2D}. The purpose of this example is to test the extrapolation capabilities of the proposed technique.

The training data were obtained from finite element simulations in which the geometry of the beam, and therefore the number of nodes, were changed, thus forcing the network to make inferences about previously unseen geometries.
The state variables are similar to the previous example but reducing everything to two dimensions:
\begin{equation*}
    \mathcal S =\left\{ \bs z=(\bs q,\bs v,\bs \sigma)\in \mathbb{R}^{2}\times \mathbb{R}^{2} \times \mathbb{R}^{3} \right\}.
\end{equation*}

The dataset is composed of $N_{\text{sim}} = 38$ simulations of 2D bending cantilever beams. The geometry varies in length from 100 to 200 and in height from 25 to 50, containing the smallest beams 126 nodes and the largest ones 450 nodes. The material's hyperelastic and viscoelastic parameters are $C_{10} = 4\cdot 10^3$, $C_{01} =  10^3$, $D_1 = 2\dot 10^{-4}$, $\bar{g}_{1} = 0.2$, $\bar{g}_{2} = 0.1$, $\tau_{1} = 0.1$, $\tau_{2} = 0.2$, respectively. A variable distributed load ranging from $F = 0.02$ to  $F = 0.04$ is applied at different positions with an orientation perpendicular to the solid surface. The 38 simulations are discretized into $N_T = 50$ time increments of $\Delta t = 2 \cdot 10^{-2}$.

The chosen hyperparameters for training include a hidden dimension of $F_h = 80$ for the node and edge latent vectors, 9 passes through the processor part, and a batch size of 4. The learning rate is set to $lr = 1 \times 10^{-3}$ with decreasing order of magnitude 0.3 at epochs 100, 200, 300 and 400, over a total of $N_{\text{epoch}} = 600$ epochs. Additionally, the training noise variance is set to $\sigma_{\text{noise}} = 4 \times 10^{-5}$.

The rollout results for the 2D viscoelastic beam are presented in Fig. \ref{fig:beam2D_resutls}. In this example, we observe that the comparative improvement with respect to vanilla GNNs is not as substantial as in the previous case. However, it is worth noting that the variability of the problem decreases considerably when reducing the number of dimensions. This reduction makes it more challenging for the network to generalize and accurately understand the physics with limited data. Nevertheless, the implementation of the nodal GENERIC bias still enhances the predictions substantially.

\begin{figure}[]
    \centering 
    \begin{tikzpicture}
        \begin{axis}[
            boxplot/draw direction=y,
            ymajorgrids=true,
            height=8cm, 
            width=6cm,  
            axis x line* = bottom,
		axis y line* = left,
            ylabel={Root Mean Squared Error},
            ymode=log, 
            cycle list={{teal_blue, fill=teal_blue, fill opacity=0.3},
                        {ash_gray, fill=ash_gray, fill opacity=0.3},
                        {mustard, fill=mustard, fill opacity=0.3}}, 
            boxplot={
                draw position={1/3 + 1*floor(\plotnumofactualtype/3) + 1/3*mod(\plotnumofactualtype,3)},
                box extend=0.2 
            },      
            xtick={0.5,1.5,2.5},
            xticklabels={%
                {$\bs q$},%
                {$\dot{\bs q}$},%
                {$\bs \sigma$},%
            },
            x tick label style={
                text width=2.5cm,
                align=center
            },
        ]
        \addplot table[row sep=newline, col sep=tab, y index=0] {PLANTILLA_INTERNET/data/mse_output_beam2D.txt};
        \addplot table[row sep=newline, col sep=tab, y index=1] {PLANTILLA_INTERNET/data/mse_output_beam2D.txt};
        \addplot table[row sep=newline, col sep=tab, y index=2] {PLANTILLA_INTERNET/data/mse_output_beam2D.txt};
        
        \addplot table[row sep=newline, col sep=tab, y index=3] {PLANTILLA_INTERNET/data/mse_output_beam2D.txt};
        \addplot table[row sep=newline, col sep=tab, y index=4] {PLANTILLA_INTERNET/data/mse_output_beam2D.txt};
        \addplot table[row sep=newline, col sep=tab, y index=5] {PLANTILLA_INTERNET/data/mse_output_beam2D.txt};
        
        \addplot table[row sep=newline, col sep=tab, y index=6] {PLANTILLA_INTERNET/data/mse_output_beam2D.txt};
        \addplot table[row sep=newline, col sep=tab, y index=7] {PLANTILLA_INTERNET/data/mse_output_beam2D.txt};
        \addplot table[row sep=newline, col sep=tab, y index=8] {PLANTILLA_INTERNET/data/mse_output_beam2D.txt};
        \end{axis}
        \begin{axis}[
            boxplot/draw direction=y,
            ymajorgrids=true,
            height=8cm, 
            width=6cm,  
            axis x line* = bottom,
		axis y line* = left,
            ylabel={Relative Root Mean Squared Error},
            ymode=log, 
            cycle list={{teal_blue, fill=teal_blue, fill opacity=0.3},
                        {ash_gray, fill=ash_gray, fill opacity=0.3},
                        {mustard, fill=mustard, fill opacity=0.3}}, 
            boxplot={
                draw position={1/3 + 1*floor(\plotnumofactualtype/3) + 1/3*mod(\plotnumofactualtype,3)},
                box extend=0.2 
            },      
            xtick={0.5,1.5,2.5},
            xticklabels={%
                {$\bs q$},%
                {$\dot{\bs q}$},%
                {$\bs \sigma$},%
            },
            x tick label style={
                text width=2.5cm,
                align=center
            },
            at={(8cm,0)}, 
        ]
        \addplot table[row sep=newline, col sep=tab, y index=0] {PLANTILLA_INTERNET/data/mse_inf_output_beam2D.txt};
        \addplot table[row sep=newline, col sep=tab, y index=1] {PLANTILLA_INTERNET/data/mse_inf_output_beam2D.txt};
        \addplot table[row sep=newline, col sep=tab, y index=2] {PLANTILLA_INTERNET/data/mse_inf_output_beam2D.txt};
        
        \addplot table[row sep=newline, col sep=tab, y index=3] {PLANTILLA_INTERNET/data/mse_inf_output_beam2D.txt};
        \addplot table[row sep=newline, col sep=tab, y index=4] {PLANTILLA_INTERNET/data/mse_inf_output_beam2D.txt};
        \addplot table[row sep=newline, col sep=tab, y index=5] {PLANTILLA_INTERNET/data/mse_inf_output_beam2D.txt};
        
        \addplot table[row sep=newline, col sep=tab, y index=6] {PLANTILLA_INTERNET/data/mse_inf_output_beam2D.txt};
        \addplot table[row sep=newline, col sep=tab, y index=7] {PLANTILLA_INTERNET/data/mse_inf_output_beam2D.txt};
        \addplot table[row sep=newline, col sep=tab, y index=8] {PLANTILLA_INTERNET/data/mse_inf_output_beam2D.txt};
        \end{axis}
        \begin{axis}[hide axis,
            xmin=0, xmax=1, ymin=0, ymax=1,
            legend style={
            at={(0.8,-0.2)},  
            anchor=north,     
            legend columns=3, 
            /tikz/every node/.append style={align=center} 
            }           
        ]
            \addlegendimage{area legend, teal_blue, fill=teal_blue, fill opacity=0.2, only marks, mark=square*}
            \addlegendentry{SPNN}
            \addlegendimage{area legend, ash_gray, fill=ash_gray, fill opacity=0.2, only marks, mark=square*}
            \addlegendentry{GNN}
            \addlegendimage{area legend, mustard, fill=mustard, fill opacity=0.2, only marks, mark=square*}
            \addlegendentry{TIGNN}
        \end{axis}
    \end{tikzpicture}
    \caption{
    Zero-shot performance for 7 different trajectories. RMSE and RRMSE values for position ($ \bs q$), velocity $ \dot{\bs q}$ and von Mises stress (\(\bs \sigma\)). Each point on the plot represents the error an individual trajectory.
    }
    \label{fig:beam2D_resutls}
\end{figure}

To evaluate the computational cost during both training and inference, we refer to Table \ref{tab:beam2D_computationalcost}. The results presented in the table align with the conclusions drawn from the previous example, highlighting the relationship between model complexity and computational efficiency. For comparison, the same simulation performed in Abaqus took 16 seconds.

\begin{table}[h]
    \centering
    \begin{tabular}{|c|c|c|c|c|}
        \hline
        & SPNN& GNN&  Nodal TIGNN &Global TIGNN\\
        \hline
        Training time per epoch (s)& 7& 10& 13&47\\
        \hline
        Inference time (s)& 0.175& 0.237& 0.306&1.15\\\hline
    \end{tabular}
    \vspace{0.3cm}
    \caption{
      Comparison of computational times for the three models. The table shows the training time required per epoch and the inference time for 50 frames. The results highlight the computational efficiency of the Nodal-TIGNN model compared to the Global-TIGNN implementation, while still outperforming the SPNN and GNN in terms of generalization and accuracy.
    }
    \label{tab:beam2D_computationalcost}
\end{table}

\subsubsection{Strong out-of-distribution generalization}

To evaluate the robustness and generalization capabilities of the trained network, we conducted an additional experiment. While the dataset primarily consists of cantilever beam examples used for training and validation, we investigated the network's ability to predict the deflection of a beam clamped at both ends---a scenario it had never encountered during training.

\begin{figure}[h]
    \centering
    \includegraphics[width=\linewidth]{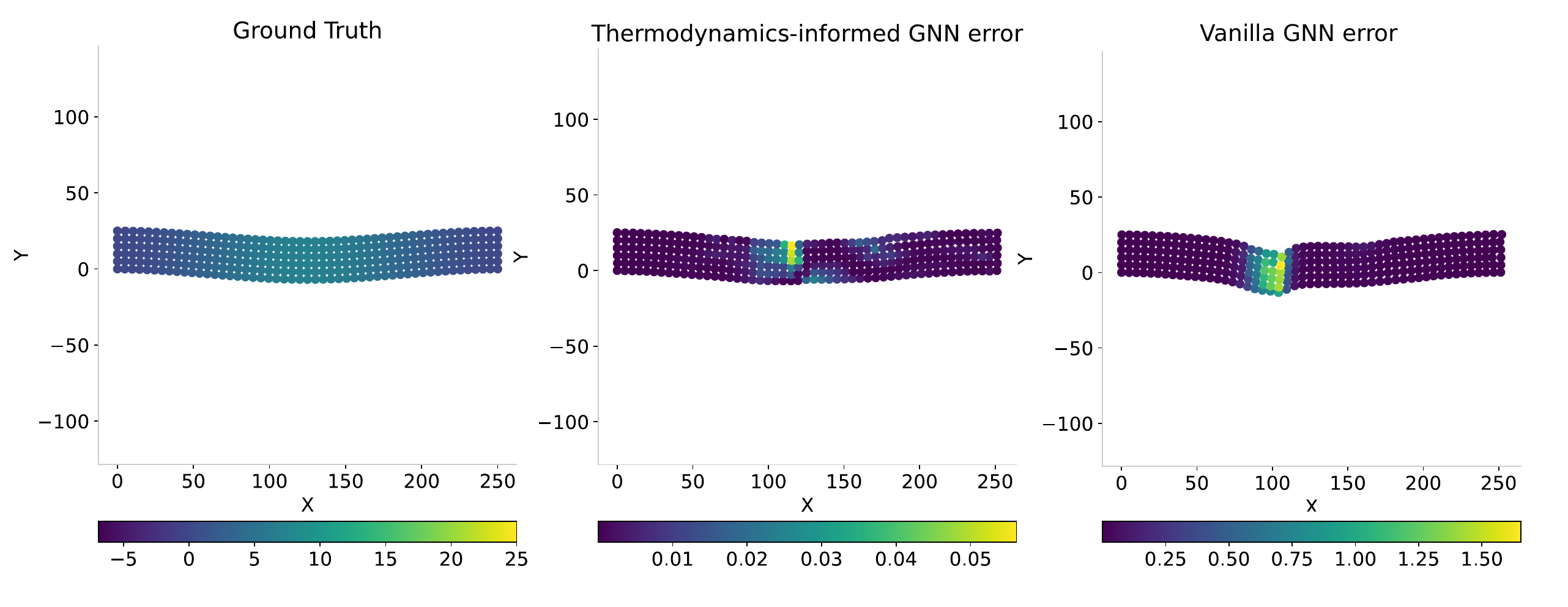}
    \caption{Last frame of the test rollout with a 2D viscoelastic beam clamped on both sides. This constitutes a scenario never seen by the network during training. Left: ground truth displacement along $y$ axis. Center: Relative error obtained with our approach. Right: Relative error in a vanilla GNN prediction for the same problem. Note the different scale of the color bar. The error is computed according to Eq. (\ref{eq:rrmse_inf}).}
    \label{fig:beam2D_2emp}
\end{figure}

Noteworthy, the geometry of the beam is set twice as large as the beams employed in the training and validation sets, resulting in a significantly higher number of nodes and a completely different geometry and boundary conditions. Consequently, in Fig. \ref{fig:beam2D_2emp}, it can be observed how well the network has generalized when exposed to a previously unseen problem.

\subsection{Liquid Sloshing}
\label{subsec: glass}

For the final example, we chose to explore fluid sloshing as a proof of concept for this nodal formulation. This selection is motivated by the fact that sloshing encompasses several challenging characteristics, including its non-linear nature and dissipative behaviour.

In this approach, we model Newtonian fluids at a hydrodynamic level under laminar conditions. At this stage, the requisite state variables for a comprehensive fluid description include position $\bs q_{j}$, velocity $\bs v_{j}$, and internal energy $ e_{j}$, as per the GENERIC formalism \cite{PhysRevLett.83.4542}. Thus, the set $\mathcal S$ of state variables is defined accordingly
\begin{equation*}
   \mathcal S =\left\{ \bs z=(\bs q,\bs v,E)\in \mathbb{R}^{3}\times \mathbb{R}^{3} \times \mathbb{R} \right\}.
\end{equation*}

The consideration of non-Newtonian fluids would need to include stresses in the state variables so as to adequately describe the energy evolution of the system \cite{moya2022physics}. 

In this case, synthetic data were obtained by Smooth Particle Hydrodynamics \cite{Monaghan_SPH}. In these simulations, the geometry of the glass is not a parameter of the problem; rather, it is filled with varying volumes. We conduct simulations with the same fluid, each with different initial velocities to initiate sloshing. In the test simulations, we evaluate the model's performance with a fluid volume (i.e., a number of mesh nodes) that has not been encountered previously.

It should also be highlighted that, unlike the previous examples, which involved solids, here we are working with a fluid in motion. This implies that the connectivity between nodes/particles changes at every time step. To calculate the new connectivity for the next time step, we consider a connectivity radius that creates new connections based on the proximity between nodes.

For the problem, we have chosen as characteristic variables a wave speed of \( c = 10 \) and a density of \( \rho = 983.2 \), while the shear viscosity is chosen as \( 1.3 \times 10^{-3} \). The dimensions of the glass are as follows: a height of \( H = 0.08 \) and a radius of \( R = 0.03 \). The glass is filled from a height of \( H_{\text{liq}} = 0.03 \) to \( H_{\text{liq}} = 0.06 \), with a variable number of particles (9000 on average). The glass is subjected to an impulsive initial force that varies from \( f_{\text{ini}} = 0.25 \) to \( f_{\text{ini}} = 0.35 \) and the connectivity radious is set to $r_c = 0.007$. For training and inference, only the fluid inside the glass has been taken into account.

The chosen hyperparameters for training include a hidden dimension of $F_h = 100$ for the node and edge latent vectors, 5 passes through the processor part, and a batch size of 1. The learning rate is set to $lr = 8 \times 10^{-4}$ with decreasing order of magnitude at epochs 100, 150, 200 and 300, over a total of $N_{\text{epoch}} = 600$ epochs. Additionally, the training noise variance is set to $\sigma_{\text{noise}} = 4 \times 10^{-5}$.

The results of four test simulations of 50 time increments each can be seen in Fig. \ref{fig:water_resutls}. In this scenario, the inclusion of the nodal GENERIC bias also improves the metrics, providing thermodynamic robustness.

\begin{figure}[]
    \centering 
    \begin{tikzpicture}
        \begin{axis}[
            boxplot/draw direction=y,
            ymajorgrids=true,
            height=8cm, 
            width=6cm,  
            axis x line* = bottom,
		axis y line* = left,
            ylabel={Root Mean Squared Error},
            ymode=log, 
            cycle list={{teal_blue, fill=teal_blue, fill opacity=0.3},
                        {ash_gray, fill=ash_gray, fill opacity=0.3},
                        {mustard, fill=mustard, fill opacity=0.3}}, 
            boxplot={
                draw position={1/3 + 1*floor(\plotnumofactualtype/3) + 1/3*mod(\plotnumofactualtype,3)},
                box extend=0.2 
            },      
            xtick={0.5,1.5,2.5},
            xticklabels={%
                {$\bs q$},%
                {$\dot{\bs q}$},%
                {$e$},%
            },
            x tick label style={
                text width=2.5cm,
                align=center
            },
        ]
        \addplot table[row sep=newline, col sep=tab, y index=0] {PLANTILLA_INTERNET/data/mse_output_water3D.txt};
        \addplot table[row sep=newline, col sep=tab, y index=1] {PLANTILLA_INTERNET/data/mse_output_water3D.txt};
        \addplot table[row sep=newline, col sep=tab, y index=2] {PLANTILLA_INTERNET/data/mse_output_water3D.txt};
        
        \addplot table[row sep=newline, col sep=tab, y index=3] {PLANTILLA_INTERNET/data/mse_output_water3D.txt};  
        \addplot table[row sep=newline, col sep=tab, y index=4] {PLANTILLA_INTERNET/data/mse_output_water3D.txt};
        \addplot table[row sep=newline, col sep=tab, y index=5] {PLANTILLA_INTERNET/data/mse_output_water3D.txt};
        
        \addplot table[row sep=newline, col sep=tab, y index=6] {PLANTILLA_INTERNET/data/mse_output_water3D.txt};
        \addplot table[row sep=newline, col sep=tab, y index=7] {PLANTILLA_INTERNET/data/mse_output_water3D.txt};
        \addplot table[row sep=newline, col sep=tab, y index=8] {PLANTILLA_INTERNET/data/mse_output_water3D.txt};
        \end{axis}
        \begin{axis}[
            boxplot/draw direction=y,
            ymajorgrids=true,
            height=8cm, 
            width=6cm,  
            axis x line* = bottom,
		axis y line* = left,
            ylabel={Relative Root Mean Squared Error},
            ymode=log, 
            cycle list={{teal_blue, fill=teal_blue, fill opacity=0.3},
                        {ash_gray, fill=ash_gray, fill opacity=0.3},
                        {mustard, fill=mustard, fill opacity=0.3}}, 
            boxplot={
                draw position={1/3 + 1*floor(\plotnumofactualtype/3) + 1/3*mod(\plotnumofactualtype,3)},
                box extend=0.2 
            },      
            xtick={0.5,1.5,2.5},
            xticklabels={%
                {$\bs q$},%
                {$\dot{\bs q}$},%
                {$e$},%
            },
            x tick label style={
                text width=2.5cm,
                align=center
            },
            at={(8cm,0)}, 
        ]
        \addplot table[row sep=newline, col sep=tab, y index=0] {PLANTILLA_INTERNET/data/mse_inf_output_water3D.txt};
        \addplot table[row sep=newline, col sep=tab, y index=1] {PLANTILLA_INTERNET/data/mse_inf_output_water3D.txt};
        \addplot table[row sep=newline, col sep=tab, y index=2] {PLANTILLA_INTERNET/data/mse_inf_output_water3D.txt};
        
        \addplot table[row sep=newline, col sep=tab, y index=3] {PLANTILLA_INTERNET/data/mse_inf_output_water3D.txt};
        \addplot table[row sep=newline, col sep=tab, y index=4] {PLANTILLA_INTERNET/data/mse_inf_output_water3D.txt};
        \addplot table[row sep=newline, col sep=tab, y index=5] {PLANTILLA_INTERNET/data/mse_inf_output_water3D.txt};
        
        \addplot table[row sep=newline, col sep=tab, y index=6] {PLANTILLA_INTERNET/data/mse_inf_output_water3D.txt};
        \addplot table[row sep=newline, col sep=tab, y index=7] {PLANTILLA_INTERNET/data/mse_inf_output_water3D.txt};
        \addplot table[row sep=newline, col sep=tab, y index=8] {PLANTILLA_INTERNET/data/mse_inf_output_water3D.txt};
        \end{axis}
        \begin{axis}[hide axis,
            xmin=0, xmax=1, ymin=0, ymax=1,
            legend style={
            at={(0.8,-0.2)},  
            anchor=north,     
            legend columns=3, 
            /tikz/every node/.append style={align=center} 
            }           
        ]
            \addlegendimage{area legend, teal_blue, fill=teal_blue, fill opacity=0.2, only marks, mark=square*}
            \addlegendentry{SPNN}
            \addlegendimage{area legend, ash_gray, fill=ash_gray, fill opacity=0.2, only marks, mark=square*}
            \addlegendentry{GNN}
            \addlegendimage{area legend, mustard, fill=mustard, fill opacity=0.2, only marks, mark=square*}
            \addlegendentry{TIGNN}
        \end{axis}
    \end{tikzpicture}
    \caption{
    Zero-shot performance for four different trajectories. RMSE and RRMSE values for position ($ \bs q$), velocity $ \dot{\bs q}$ and internal energy ($e$).
    }
    \label{fig:water_resutls}
\end{figure}

\begin{figure}[h!]
    \centering
    \includegraphics[width=\linewidth]{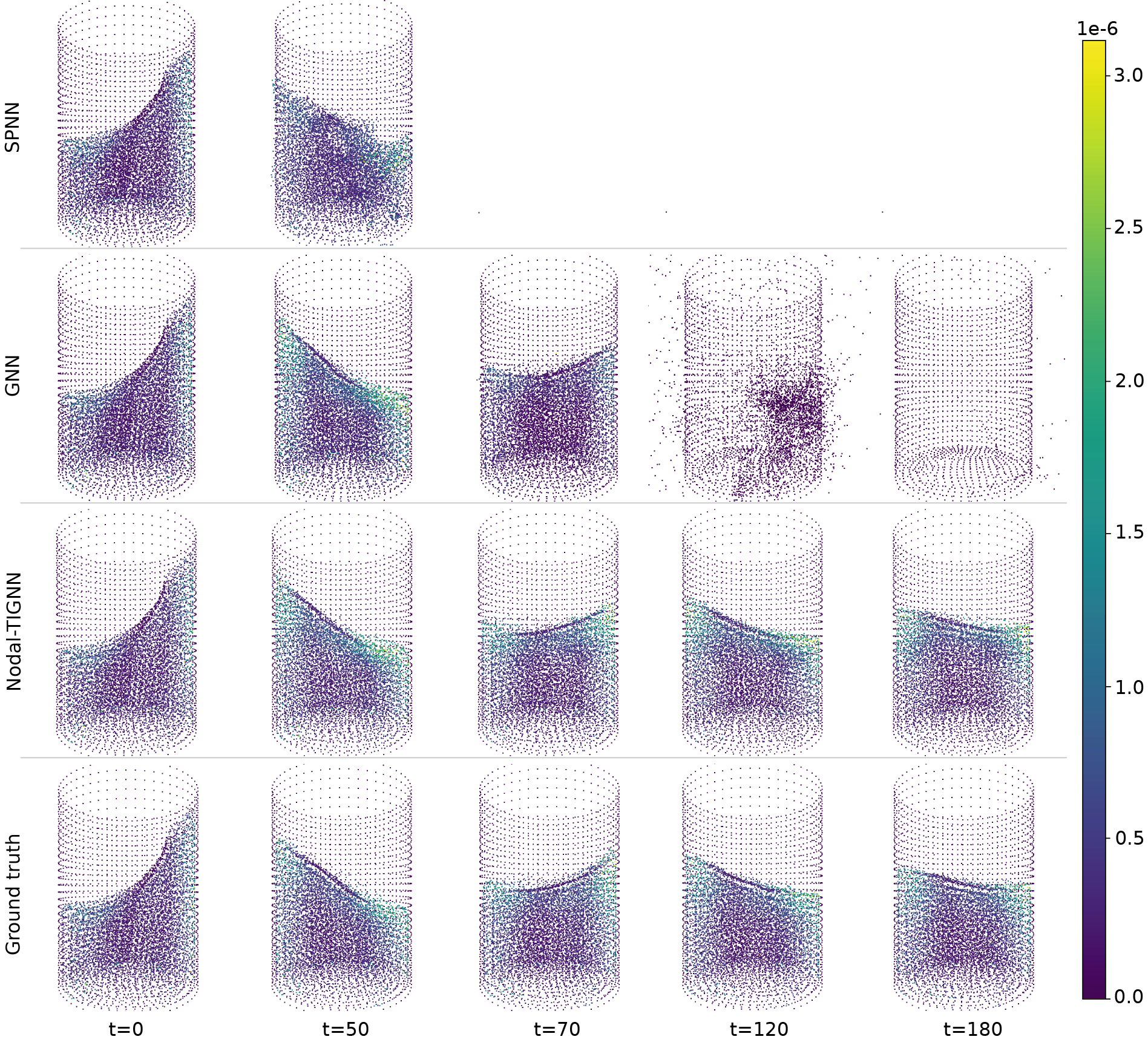}
    \caption{
    Results for a 200-time-increment simulation of sloshing in a glass of water. Five snapshots of the sloshing sequence were selected for comparison, corresponding to time increments number 10, 50, 70, 120, and 180 from left to right. The first row shows the results of the prediction using the SPNN network, which fails after step 64. The second row presents the results of inference using the GNN network across all selected frames. The third row corresponds to the fluid reconstruction from our proposed method, and the fourth row shows the ground truth, SPH simulation. The color of the particles in all rows indicates the energy level of each node. }
    \label{fig:water_roll}
\end{figure}

 In Fig. \ref{fig:water_roll}, we present a rollout of an unseen simulation with a number of nodes that the network has never encountered before, spanning 200 time steps---four times more than the 50 time steps employed for training, thus demonstrating the strong generalization ability of the proposed method---. The visualization demonstrates the network's ability to learn the dissipative behaviour of the fluid, gradually bringing it to a halt. The results also illustrate how the SPNN model fails before reaching frame 70, while the GNN demonstrates slightly greater robustness, but is unable to complete the simulation successfully.

Finally, in terms of computational efficiency, Table \ref{tab:water3D_computationalcost} demonstrates that while the nodal TIGNN model incurs a higher computational cost compared to simpler architectures, it remains memory-efficient and manageable, unlike the global TIGNN model, which exceeds memory limitations.For comparison, running the same simulation in Abaqus requires approximately 74 seconds, which is significantly longer than the inference time of the presented networks. In this last case study, it is important to note that the inference time also includes the computation of graph edges, as they are recalculated at each new time step. This edge computation takes approximately 2 seconds per step, yet the overall prediction process using the proposed architectures remains substantially faster than traditional simulation methods.

\begin{table}[h]
    \centering
    \begin{tabular}{|c|c|c|c|}
        \hline
        & SPNN& GNN&  Nodal TIGNN \\
        \hline
        Training time per epoch (s)& 116& 833& 1084\\
        \hline
        Inference time (s)& 1.57& 2.99& 4.13\\\hline
    \end{tabular}
    \vspace{0.3cm}
    \caption{
      Comparison of computational times for the three models. The table shows the training time required per epoch and the inference time for 50 frames. This table does not show the computation time of the global-GNN model, because it requirements were so large that it did  not fit in memory.
    }
    \label{tab:water3D_computationalcost}
\end{table}

\section{Conclusions}
\label{sec:conclusions}

The introduction of inductive biases in the improvement of the learning process of physical phenomena has brought substantial improvements in terms of accuracy and robustness of predictions. Combined with graph neural networks, these inductive biases provide results with at least an order of magnitude higher accuracy. However, the global form of these thermodynamic biases destroys the nodal structure of graph networks, which makes them so attractive.

 In this work we have developed a local form for the metriplectic (GENERIC) bias that ensures the thermodynamic correctness of the predictions at inference and improves the accuracy obtained by black-box approaches. Firstly, by redesigning the implementation of GENERIC at the nodal level, we ensure physical consistency. Secondly, leveraging a graph neural network-based architecture enables us to exploit the inherent geometric structure of the problem. Notably, our nodal implementation of GENERIC eliminates the need for assembling $\bs L$ and $\bs M$ matrices, a challenging task in systems comprising tens of thousands of nodes that can be impractical in certain computational environments. Consequently, we maintain the efficiency of GNNs while overcoming computational bottlenecks. The water sloshing problem, for instance, crashed while run under the previous, global implementation of the GENERIC bias when running on a PC equipped with an Nvidia GeForce RTX 3090 GPU due to lack of memory space. Other than that, the proposed method achieves levels of accuracy similar to that obtained in our previous implementations, see \cite{hernandez2022thermodynamics}.

Moreover, our method demonstrates significant computational efficiency compared to traditional finite element methods, accelerating processing time by at least one order of magnitude, in the worst case scenario encountered during this work. This efficiency gain underscores the practical applicability and effectiveness of our approach. Additionally, our method exhibits strong generalization capabilities by making accurate inferences on examples significantly different from those encountered during training, as demonstrated by the examples with beams clamped at both ends.

Looking ahead, future work will explore training with different materials within the same session. For instance, investigating the behavior of Newtonian and non-Newtonian fluids and exploring variations in geometry to a greater extent.

\section*{Acknowledgements}

This work was supported by the Spanish Ministry of Science and Innovation, AEI/10.13039/501100011033, through Grant number TED2021-130105B-I00 and by the Ministry for Digital Transformation and the Civil Service, through the ENIA 2022 Chairs for the creation of university-industry chairs in AI, through Grant TSI-100930-2023-1.

This research is also part of the DesCartes programme and is supported by the National
Research Foundation, Prime Minister Office, Singapore under its Campus for Research
Excellence and Technological Enterprise (CREATE) programme.

This material is also based upon work supported in part by the Army Research Laboratory and the Army Research Office under contract/grant number W911NF2210271.

The authors also acknowledge the support of ESI Group through the chairs at the
University of Zaragoza and at ENSAM Institute of Technology.



\end{document}